\documentclass{paper}
\usepackage{setspace}
\usepackage{amsmath}
\usepackage{amssymb}
\usepackage{amsthm}
\usepackage{hyperref}
\usepackage{graphicx}
\usepackage{mathrsfs}
\usepackage{comment}
\usepackage[top=1.25in, bottom=1.25in, left=1.25in, right=1.25in]{geometry}

\usepackage{amsthm}

\newtheorem{thm}{Theorem}

\newtheorem{prop}[thm]{Rule} 

\newtheorem{biWe knowb}[thm]{Reference}
\theoremstyle{definition}

\title{Binocular Disparity as an Explanation for the Moon Illusion}
\author{Joseph Antonides\textsuperscript{1} and Toshiro Kubota\textsuperscript{2}}

\begin{document}
\maketitle

\raggedright
{\small 1. {\it Corresponding author}. The Ohio State University Department of Mathematics, 231 W. 18th Avenue, Columbus, OH, USA 43210. E-mail: antonides.4@osu.edu. Phone: (614) 247-4717.\\
2. Susquehanna University Department of Mathematical Sciences, Selinsgrove, PA 17870. E-mail: kubota@susqu.edu.\\}

\raggedright

\begin{abstract}
We present another explanation for the moon illusion, the phenomenon in which the moon looks larger near the horizon than near the zenith. In our model of the moon illusion, the sky is considered a spatially-contiguous and geometrically-smooth surface. When an object such as the moon breaks the contiguity of the surface, instead of perceiving the object as appearing through a hole in the surface, humans perceive an occlusion of the surface. Binocular vision dictates that the moon is distant, but this perception model contradicts our binocular vision, dictating that the moon is closer than the sky. To resolve the contradiction, the brain distorts the projections of the moon to increase the binocular disparity, which results in an increase in the perceived size of the moon. The degree of distortion depends upon the apparent distance to the sky, which is influenced by the surrounding objects and the condition of the sky. As the apparent distance to the sky decreases, the illusion becomes stronger. At the horizon, apparent distance to the sky is minimal, whereas at the zenith, few distance cues are present, causing difficulty with distance estimation and weakening the illusion.
\end{abstract}

\vspace{.1in}


\section*{Introduction}
The moon illusion has puzzled scientists for centuries - why does the moon appear to be larger at the horizon than at higher elevations in the sky? The illusory effect is not unique to the moon; in fact, the illusion was once known as the ``celestial illusion," for the sun, constellations, and many other celestial objects look larger near the horizon than higher in the sky as well.\textsuperscript{1} 
How do we know this to be an illusory phenomenon and not just a physical occurrence of nature? One can perform a simple demonstration of the illusion by taking photographs that illustrate the moon's path across the night sky. In these photographs, the moon subtends a constant angular size of approximately 0.52 degrees.\textsuperscript{2} 
According to Hershenson, written records extending back to the seventh century B.C.E. in the cuneiform script of ancient Samaria reference the moon illusion,\textsuperscript{3} 
and scientists have proposed dozens of theories since the time of Aristotle in the fourth century B.C.E.,\textsuperscript{4} 
but no single theory has been accepted by the scientific community at large.  \\

\vspace{.1in}

This paper serves two purposes: (i) to briefly introduce two classic theories of the moon illusion in order to provide the reader with some context, and (ii) to present a new theory that suggests the moon illusion is fundamentally caused by a contradiction between binocular cues and occlusion cues due to perception. 
\vspace{.1in}

\section{Apparent-Distance Theory and Size-Contrast Theory}
Scientists have attempted to solve the riddle of the moon illusion for centuries. Aristotle himself theorized that the atmosphere of the Earth magnified the moon, a theory no longer considered a possibility; again, this is an illusory effect, not a physical occurrence of nature. Because the scientific community has yet to accept one ubiquitous explanation, many theories currently exist that try to explain the moon illusion. Two of the oldest (and perhaps deserving of the term ``classic") theories are the Apparent-Distance theory and the Size-Distance theory.  \\

\vspace{.1in}

The Apparent-Distance theory was first proposed by the 11th century mathematician Ibn al-Haytham, otherwise known as Alhazen. The theory was popularized in 1962 by Kaufman and Rock.\textsuperscript{5}
At the time, Kaufman and Rock's argument was particularly influential, making its way into textbooks as the general explanation for the moon illusion. Essentially, the Apparent-Distance theory claims that humans perceive the sky as a two-dimensional plane. As objects move closer to the horizon, the perceived distance to the object increases. One version of the Apparent-Distance theory utilizes Emmert's Law, which states that perceived size of an object is proportional to the perceived distance to the object. Another version of the Apparent-Distance theory assumes an inherent anisotropy of visual space, in which humans underestimate distances in the vertical direction as compared with the horizontal direction.\textsuperscript{6} 
The Apparent-Distance theory is frequently illustrated using the ``flattened sky dome" model as shown in Figure 1.\\

\vspace{.1in}


\center \includegraphics[scale=.07]{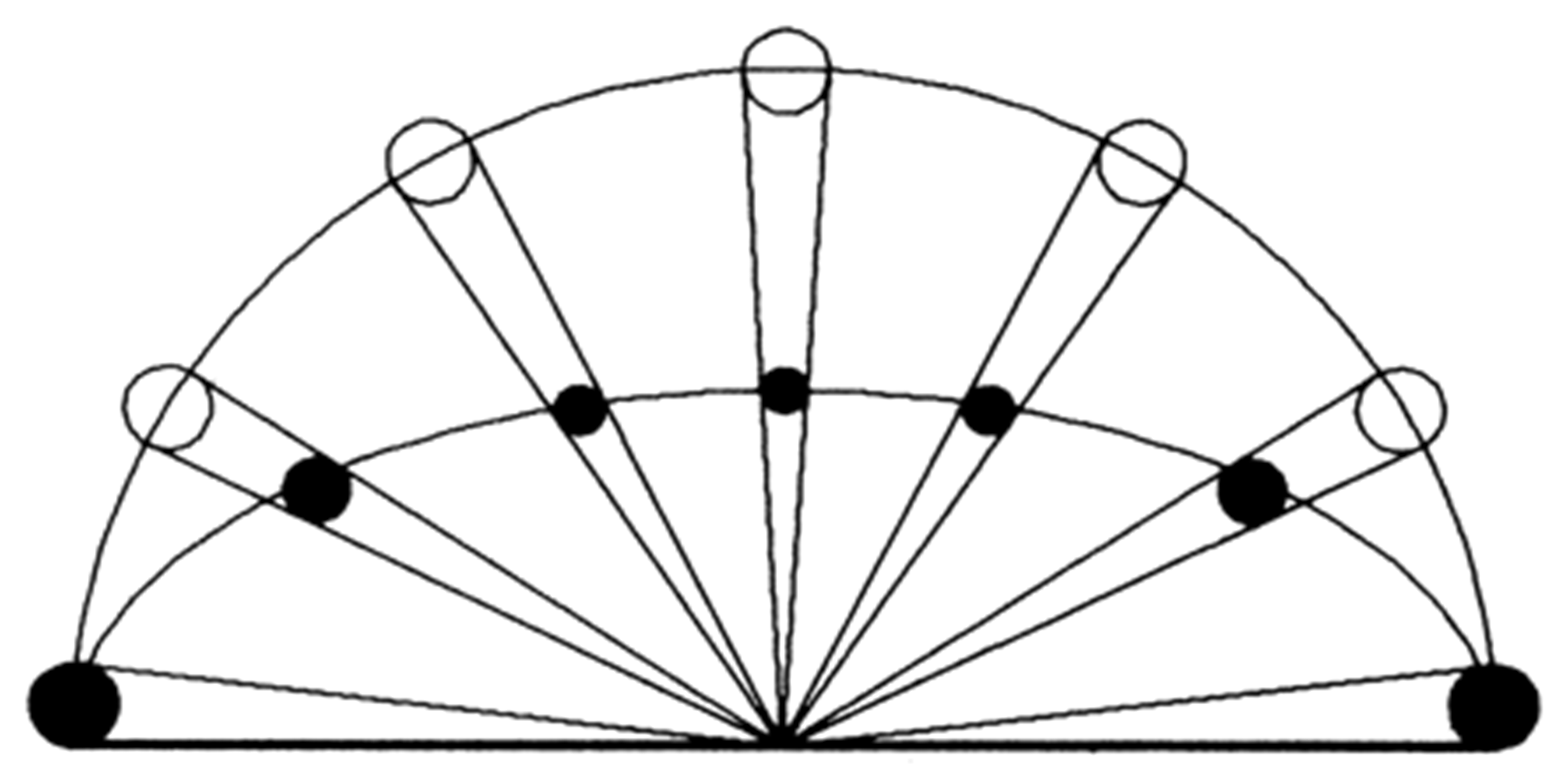}

\vspace{.05in} {\small {\bf Figure 1.} The ``Flattened Sky Dome" illustrates the\\ Apparent-Distance theory of the moon illusion}

\raggedright

\vspace{.1in}

Again, this theory states that as objects move closer to the horizon, they appear smaller. Therefore, for an object to appear larger at the horizon, the object would also have to appear further away. However, most observers who experience the moon illusion claim that the moon appears closer, not further, creating a size-distance paradox. Advocates of the Apparent-Distance theory have difficulty explaining this contradiction. Also, according to Kaufman and Rock,\textsuperscript{7}
the visible terrain is essential for the moon illusion to occur. However, experimentation by Suzuki\textsuperscript{8}
makes evident that the illusion can be experienced  with no visible terrain. \\

\vspace{.1in}

The most common alternative to the Apparent-Distance theory is the Size-Contrast theory, a theory developed by scientists such as Restle\textsuperscript{9} 
and more recently by Baird, Wagner, and Fuld.\textsuperscript{10}
The Size-Contrast theory suggests that the perceived size of the moon is proportional to the visual  size of a referent object. In other words, the Size-Contrast theory suggests that the perceived size of the moon is affected the sizes of the objects surrounding the moon. At the horizon, the objects surrounding the moon include trees, buildings, mountains, etc., all of which are on our plane of existence, and we have a perceived notion of the sizes of these objects. The horizon moon, being comparable in size to the sizes of these familiar objects, appears relatively large. The moon at higher elevations is compared to the expansive night sky and surrounding clouds, resulting in a smaller-appearing moon. The Size-Contrast theory can be likened to the Ebbinghaus illusion (Figure 2). The perceived sizes of the center circles are affected by the sizes of their respective context circles. \\

\vspace{.1in}


\center 

\includegraphics[scale=.03]{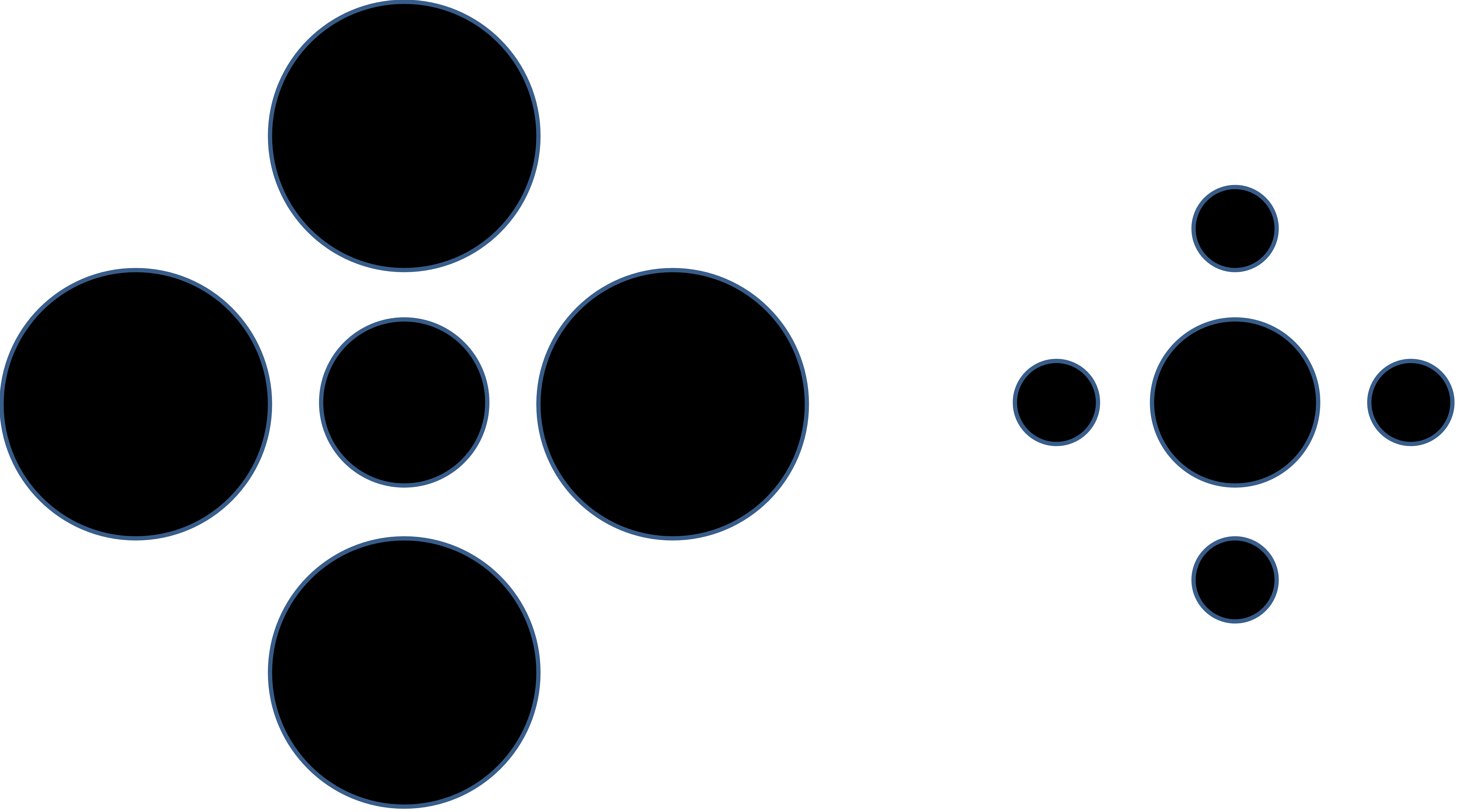}

\vspace{.05in} 

{\small {\bf Figure 2.} The ``Ebbinghaus illusion" emphasizes the effect of \\ angular sizes of referent objects to the perceived size of an object}

\raggedright

\vspace{.1in}

The reader should note that these are but two (rather old) theories to explain the moon illusion, provided to give some context to the reader who may not be so well-read about the moon illusion. For a survey of (especially) more modern theories of the moon illuion, the reader is encouraged to reference Hershenson.\textsuperscript{3} 
Hershenson provides a survey of theories by Enright, Roscoe, Gogel and Mertz, Leibowitz, and many others. The reader is also encouraged to reference Ross and Plug\textsuperscript{11} 
who provide an elegant history of the moon illusion in their text {\it The Mystery of The Moon Illusion}.

\section{Our Proposal}
We propose a new theory to explain the moon illusion. In general, visual illusions frequently manifest how humans infer a three-dimensional world from two-dimensional projections. These illusions arise when the inference fails to predict the reality. Such illusions include the Charlie Chaplin ``hollow mask" illusion,\textsuperscript{12} 
the Checkershadow illusion,\textsuperscript{13} 
and the Barberpole illusion.\textsuperscript{14} 
We believe the moon illusion is no exception. Our goal is to explain the moon illusion from a causal perspective; what inferential rules employed by our brain can cause a problem in viewing of the moon, resulting in the moon illusion? \\

\vspace{.1in}

We propose the establishment of two rules for our visual perception:\\

\begin{prop}\label{prop1} Humans perceive spatially homogeneous areas as spatially-contiguous surfaces.\end{prop}

\begin{prop}\label{prop2} Humans perceive small areas disturbing homogeneous areas as objects occluding the surface, rather than appearing through a hole in the surface. \end{prop} 

\vspace{.1in}

We derive these rules of visual perception from Donald Hoffman's rule of generic views, which states that humans ``[c]onstruct only those visual worlds for which the image is a stable (i.e., generic) view" (Hoffman,\textsuperscript{15} 
p. 25). Applying Hoffman's rule of generic views to our first rule of visual perception, we discover that if a homogeneous area was not interpreted as a contiguous surface, then the homogeneity of the area would be disturbed by a gap simply by rotating one's head. Applying Hoffman's rule to our second rule of visual perception, we discover that objects appearing through a hole in a surface would be unstable against changes in viewpoint, as it would be subject to occlusion by the surface. \\

\vspace{.1in}

The sky, being a homogeneous area, is interpreted as a spatially-contiguous surface (Rule 1). Consequently, the moon, disturbing the homogeneous area, is interpreted as occluding the surface of the sky rather than appearing through a hole in the surface (Rule 2) and is therefore perceived as being closer. However, binocular vision dictates that the moon is distant. This poses a dilemma: perception does not agree with binocular vision. \\

\vspace{.1in}

The brain distrusts binocular disparity; after all, sensory signals are noisy and unpredictable.\textsuperscript{16} 
Foley\textsuperscript{17}
has demonstrated that binocular disparity is not veridical.\textsuperscript{16} He presents a relation he calls the ``effective retinal-disparity invariance hypothesis." Simply put, the relation states that the distance to one point, a reference point $r$, is determined by an egocentric distance signal, a signal that dominates perception and allows perception to gauge distance to an object at point $r$. This distance signal is commonly misperceived because ``far" objects near this egocentric distance signal frequently appear nearer than they actually are, and ``near" objects frequently appear farther than they really are. \\

\vspace{.1in}

To solve this dilemma, the brain distorts the projections of the moon, causing the perceived angular size of the moon to expand. The degree of distortion is dependent upon the perceived distance to the sky. In reference to Foley's model, we consider the sky to be the provider of an egocentric distance signal to the moon. The egocentric distance signal is influenced by objects called distance cues, which alter our perception's perceived distance. (Consequently, the perceived distance to the sky is influenced by these distance cues that may or may not be available, depending on the particular location of the moon.) For example, at the horizon, distance cues such as mountains, trees, or buildings are usually available. These objects that exist on our plane of existence provide our perception with an estimation of the distance to the sky. At higher elevations, distance cues are not as readily available. As a result, our perception has difficulty estimating the distance to the sky; the ``sky dome" is indeterminate at high elevations. The degree of distortion is low, resulting in a weakened illusion and a smaller-appearing moon. Figure 3 illustrates the effect cues to distance have on the illusory phenomenon. \\

\vspace{.1in}

\center 

\includegraphics[scale=.5]{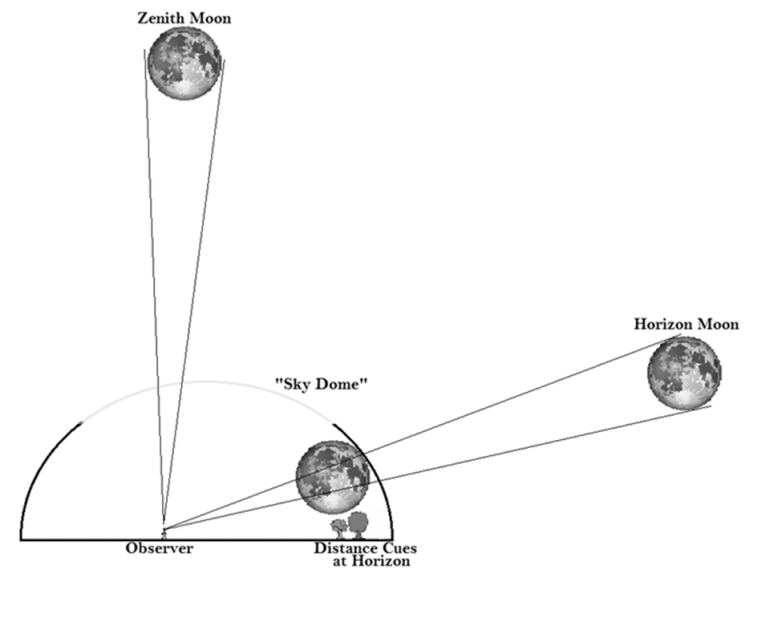}

\vspace{.05in} 

{\small {\bf Figure 3.} Distance cues such as trees affect\\ the degree of the illusory phenomenon}

\raggedright
\vspace{.1in}

We derived a function that models the angular expansion of an object due to displacement ratio:    

$$ 
\xi = 2\arctan\Big(\frac{\tan(\frac{\theta}{2})}{1 - \frac{\Delta z}{z}}\Big)
$$

where $\xi$  is the expanded angular size of the object, $\theta$ is the actual angular size of the object, $z$ is the actual distance to the object, and $\Delta z$ is the displacement of the object. Figure 4 illustrates these notations with $P$ being the actual location and $\hat{P}$ being the location after the displacement.\\

\vspace{.1in}

\center 
\includegraphics[scale=.07]{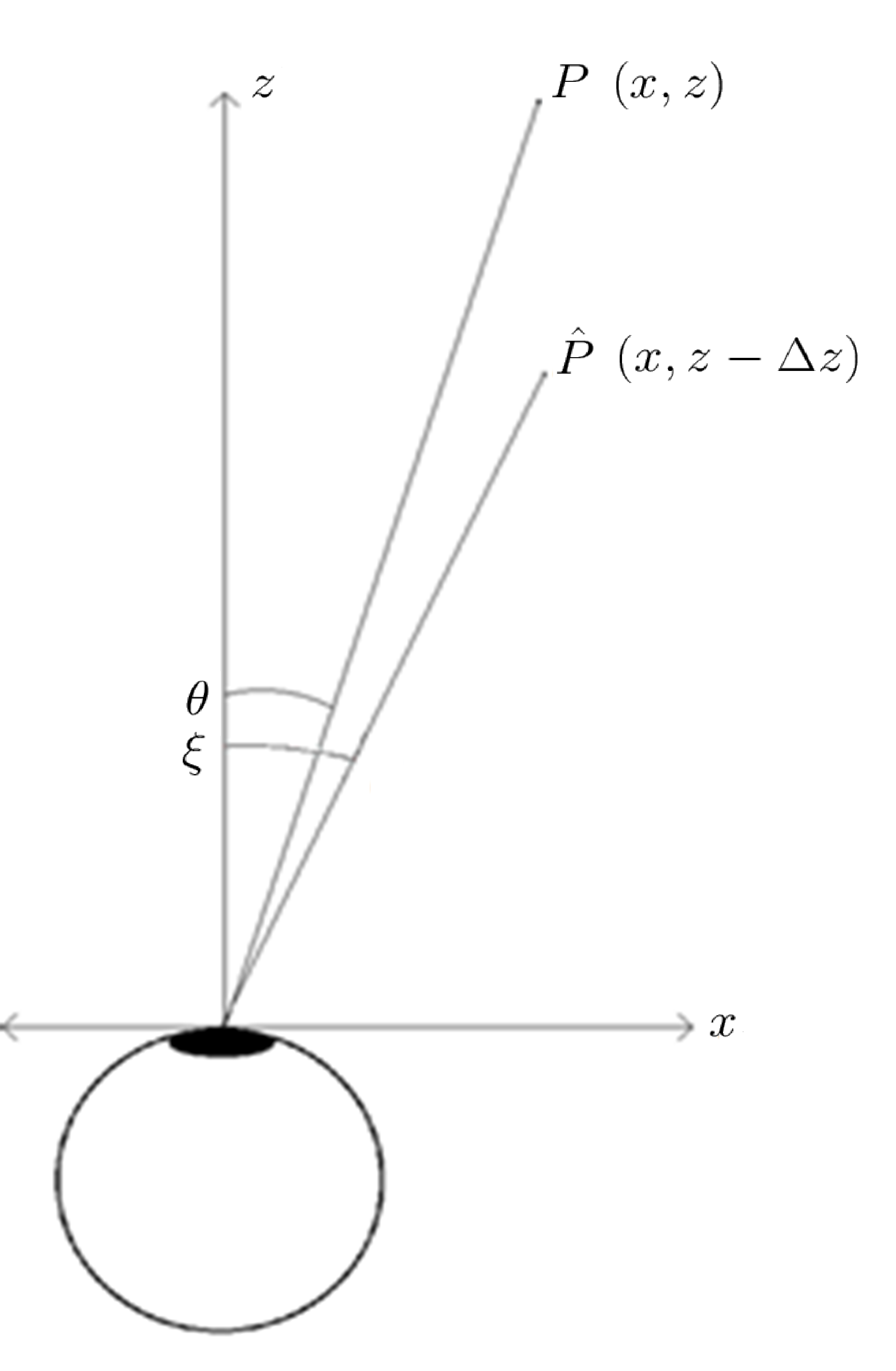}
\vspace{.05in} 

{\small {\bf Figure 4.} Disparity between actual size and perceived size\\ increases as disparity between actual distance to an\\ object and perceived distance to an object increases}

\raggedright

\vspace{.1in}

For the moon, $z \approx 384400$ km, and $\theta \approx 0.518^\circ$. Using these values, we can plot the function for $\xi$ as shown in Figure 5. The rate of the angular size increase is very small near zero displacement ratio. To achieve, say, 20\% increase in the angular size, the displacement ratio has to be close to 0.2. As such, the egocentric distance signal (the sky) must be greatly misperceived by objects near the reference point (the moon), such as mountains, trees, buildings, etc. At the zenith, little to no misperception of the distance to the sky is possible; in fact, the brain has difficulty gauging any perception of the sky at higher elevations.\\

\vspace{.1in}

\center 
\includegraphics[scale=.07]{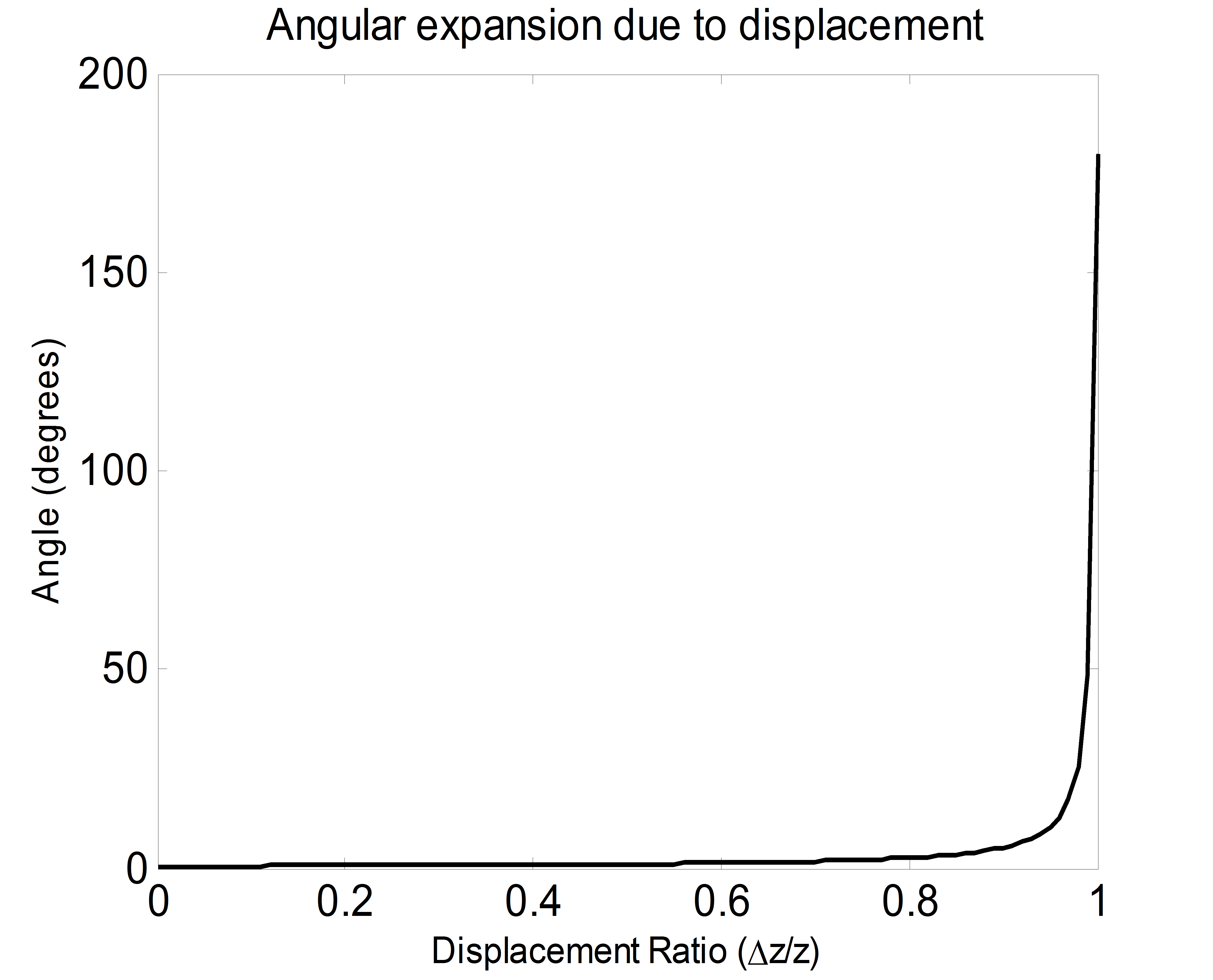}
\vspace{.05in} 

{\small {\bf Figure 5.} Plot of $\xi$, demonstrating the\\ rapid increase of angular distortion as\\ displacement ratio increases}

\raggedright

\vspace{.1in}

One famous observation is that the illusion is diminished or removed altogether when one views the moon and night sky upside down. We think this phenomenon (or lack thereof) occurs because the human perception of the world model has been distracted, and humans cannot estimate the distance to the sky. The visual system's binocular percepts may also be distracted, furthering the uncertainty as the distance to the moon itself is made incomprehensible. \\

\section{Results and Discussion}

Observation of the rise of the full moon indicated that objects in the moon's visual field also look larger. For example, clouds partially covering the moon experience the same rate of expansion when the viewing is switched from monocular to binocular. Similar observations were reported in Biard et al. This is consistent with our theory, for the distortion of the projection should affect anything along the paths of the projection by the same amount. \\

\vspace{.1in}

We also conducted a survey to examine apparent distance perception of humans. The survey was approved by the Susquehanna University Institutional Review Board (20120712.1). We used Susquehanna University's directory of students, faculty, staff, and alumni to send invitations (via e-mail) to take the survey. We invited all persons listed in the directory. Approximately 415 people participated in this survey, on a purely volunteer basis. The survey consisted of five images, with three dominant objects in each image. The possible permutations of the objects in the images were given in multiple choice format. The following instructions were given to the participants: ``From the choices provided below, please select the order in which the objects in the image appear, from closest to farthest. Note: we are testing your perception and not your spatial reasoning, so please select your answer based on your first impression of the picture." The only image of interest to us was the image of the moon; the other images were used in order to obscure our goal from the participants. Figure 6 is a collage of the five images used in the survey, all of which were taken from the Berkeley Segmentation Dataset and Benchmark.\textsuperscript{18}

\vspace{.1in}

\center 
\includegraphics[scale=.75]{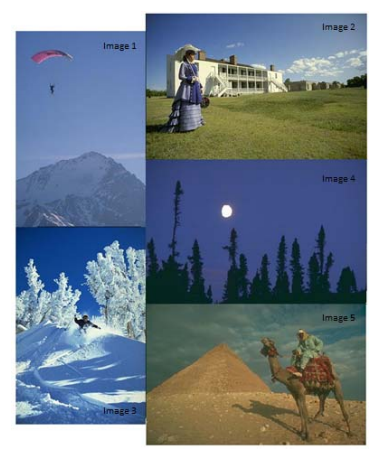}
\vspace{.05in} 

{\small {\bf Figure 6.} A collage of the images used in our survey which\\ examined humans' apparent distance perception}

\raggedright

\vspace{.1in}

Image 4 depicted the moon in the sky with trees in the background. The moon in the image was not a full moon; it was approximately 70\% illuminated. As neither the moon nor the trees occluded each other, their order with respect to each other is irrelevant to our study. Only the order of the moon and the sky with respect to each other is analyzed. Results indicated that 70.6\% of participants perceived the moon as being closer than the sky, and 29.4\% perceived the moon as being farther than the sky. Thus, 70.6\% agreed with our hypothesis that the moon is perceived as occluding the sky. However, 29.4\% of participants who did not agree with the hypothesis were more than we initially anticipated. We think that reasons for the less dominant result were that the sky in the image appears dark without much details (thus it does not display strong physical presence) and, despite the warning against it, some participants may have responded cognitively using their scientific knowledge. \\

\vspace{.1in}

In the future, we would like to artificially induce the moon illusion. The task is to construct visual stimuli, a stereo pair of computer-generated images, which induces conflicts between perception and binocular vision. The computer-generated scene consists of a moon-like object with zero binocular disparity, a sky-like plane surrounding the moon-like object, and other objects with non-zero binocular disparity that provide distance cues to the sky-like plane.  If our proposal is correct, the stimuli, when fused binocularly, may induce expansion of the moon-like object to our perception. Then, we can experiment various distance cues and conditions of the sky to determine which factors are influential to the illusion. Under such controlled environment, we can add additional perceptual cues and study their effects on the illusion. For example, do motion cues strengthen the illusion? Do fake shadows strengthen the illusion?\\

\vspace{.1in}

We would also like to quantitatively test our theory of the moon illusion by measuring the correlation between perceived distance to the sky prior to observing the rise of the moon and perceived angular size increase of the moon at the horizon, in different environments (open field, valley, mountain, etc.). Although measures are subjective, if the two are highly correlated in a way that the shorter the perceived distance is, the larger the increase of the moon size becomes, then we have supporting evidence for our theory. This will allow us to gauge the degree of perceived expansion of the moon with respect to the perceived displacement, resulting in a mapping analogous to Eq (1).

\section{Acknowledgements}
This work was supported by United States National Science Foundation grant CCF-1117439. We would also like to thank Briley Acker, Herman de Haan, and Jessica Ranck for discussion and feedback.

\bibliographystyle{apacite}

\end{document}